# Gesture Control of Micro-drone: A Lightweight-Net with Domain Randomization and Trajectory Generators

Isaac Osei Agyemang[a,*], Isaac Adjei Mensah[a], Sophyani Banaamwini Yussif [c], Fiasam Linda Delali[b], Bernard Cobinnah Mawuli[c], Bless Lord Y. Agbley[c], Collins Sey[b], and Joshua Berkoh[d]

[a] *School of Information and Communication Engineering, University of Electronic Science and Technology of China*
[b] *School of Information and Software Engineering, University of Electronic Science and Technology of China*
[c] *School of Computer Science and Engineering, University of Electronic Science and Technology of China*
[d] *School of Information Technology, University of Cincinnati*

**Abstract:** Micro-drones can be integrated into various industrial applications but are constrained by their computing power and expert pilots, a secondary challenge. This study presents a computationally-efficient deep convolutional neural network that utilizes Gabor filters and spatial separable convolutions with low computational complexities. An attention module is integrated with the model to complement the performance. Further, perception-based action space and trajectory generators are integrated with the model's predictions for intuitive navigation. The computationally-efficient model aids a human operator in controlling a micro-drone via gestures. Nearly 18% of computational resources are conserved using the NVIDIA GPU profiler during training. Using a low-cost DJI Tello drone for experiment verification, the computationally-efficient model shows promising results compared to a state-of-the-art and conventional computer vision-based technique.

## 1. Introduction

Integration of Unmanned Aerial Vehicles (UAV) in diverse application domains has induced positive results by mitigating threats associated with workers and accelerating processes [1]. UAV's usage in military, agriculture, energy, rescue operations, infrastructure inspection, and other industrial applications complements the assertion of threat mitigation and process acceleration, as mentioned earlier. UAVs are categorized under two broad perspectives, macro and micro. Within each category is either a multi-rotor, fixed-wing, single-rotor helicopter, or a fixed-wing hybrid UAV. Macro UAVs (e.g., Freefly Alta 8, DJI Matrice Series, Intel's Falcon+, etc.) come equipped with enough compute resources, hence feasible with the integration of extra sensing modalities (e.g., LIDAR, SONAR, etc.). In contrast, micro UAVs (e.g., Crazyflies, DJI Tello, etc.) are resource-constrained edge nodes; hence does not support the integration of additional sensing modalities. Control of UAVs is dominated by designated controllers embedded with navigational algorithms or flight control software running on Ground Control Station (GCS) hardware elements. Such intuitive UAV control is associated with macro UAVs. The lack of computational resources has rendered micro UAVs in industrial applications almost irrelevant. However, testing pilot-based projects and basic aerial tasks (e.g., surveillance, structure monitoring, etc.) can make use of micro UAVs to reduce the cost associated with macro UAVs. The development into virtual reality is exhibiting promising signs of control of drones via hand gestures which is a suitable navigation control for micro UAVs and an alternative for macro UAVs. Currently, hand gesture control is in three folds; (1) data-glove, (2) radar-based, and (3) vision-based [2].

Most existing drone gesture control works have used data-gloves and radar-based control approaches. In contrast, the few utilizing vision-based include guidance elements (e.g., LEAP, MAVLink Drone controller, and other wearables). In [2], a leap motion device is used to take in data and train a convolutional neural network (CNN) and a neural network that adduces accuracies of 89.6% and 98.9% to control a drone. Montebaur *et al.* [3] proposed using a wearable ring equipped with a captive sensor to send hand gesture data to a wearable computer to be processed and send navigational commands to the drone. In [4], the authors proposed three deep learning methods (CNN, VGG-16, and ResNet-50) trained on gesture data to control a drone. The author's approach is verified using Raspberry Pi, Ar Drone, and LEAP. Contrary to the deep learning methods, using wearable and a fabricated box unit, hand gestures are converted into radio signals and sent as navigational commands to a quadcopter [5]. Static and non-static hand gesture data via LEAP is used to control a synthetic drone in 3D Unity environment. The detection process focuses on finding angles related to hand gestures and classifying the gestures [6]. Dictionary learning and sparse representation are used to recognize hand gestures to control a drone [7]. The authors tested their approach using Robot Operating System (ROS) framework and an AR Drone. Haar features AdaBoost algorithm is used to recognize five hand gestures to control a Parrot AR 2.0 drone [8]. Drawbacks associated with the

---

*Corresponding author: School of Information and Communication Engineering, University of Electronic Science and Technology of China.
Email address: ioagyemang@std.uestc.edu.cn

**Excerpt of the experiment:**
https://www.youtube.com/watch?v=MxRmKbnkXOY

reiterated non-deep learning methods include using extra peripherals (data-gloves, wearables, etc.), which require calibrations and are costly. The existing hand-gesture deep learning methods have a computation bottleneck for which resource-constrained drones are deficient.

Akin to the existing hand gesture drone control, this study presents a computationally-efficient deep convolutional neural network (DCNN) hand gesture control for a low resource-constrained drone. The computationally-efficient DCNN is a variant of the MobileNet [9] dubbed G-MobNet, within which Gabor filters [10] and Spatial Separable Convolutions [11] are used to mitigate computation. Further, Convolutional Block Attention Module (CBAM) [12], perception-based action space, and trajectory generators are utilized to enhance performance. The rest of the paper ensues as follows: (1) Section 2 entails a crisp description of the proposed method, (2) Section 3 details the experiment and results, and (3) a concluding remark is drawn in Section 4.

## 2. Computationally-efficient gesture controller

This section presents the configuration of the architecture of G-MobNet, data randomization applied to the dataset and the mechanism of gesture control.

### 2.1. Gabor Filters

Gabor filters are sinusoids modulated by a Gaussian function suitable for disparity estimations, feature extraction, and texture analysis by exploiting frequencies in specified regions of images. Gabor filters have exhibited the urge to extract meaningful features which are sometimes ignored or not understood by convolutional filters [13]–[15]. Gabor filters have seen a fair integration in CNN/DCNN models with promising results [16]-[17]. Computaionalwise, kernels of Gabor filters have a computational complexity of $N \times N$ whereas kernels of standard convolutions are $(N \times N)^2$. As such, Gabor filters require fewer computing resources than convolutional filters. Through a Gabor filter bank, Gabor filters (generic filters) can be generated using a Gabor function expressed as:

$$g(x,y;\lambda,\theta,\varphi,\sigma,\gamma) = \exp\left(-\frac{x'^2 + \gamma^2 y'^2}{2\sigma^2}\right) exp\left(i\left(2\pi \frac{x'}{\lambda} + \varphi\right)\right) \quad (1)$$

where $x' = x\,cos\theta + y\,sin\theta$, $y' = -x\,sin\theta + y\,cos\theta$, and $i$ be an imaginary unit. Parameters $x, y$ are the Gabor kernel size, $\lambda$ denotes a wavelength, $\theta$ represents the orientation, $\varphi$ signifies phase offset, $\sigma$ is bandwidth, and $\gamma$ is the aspect ratio. Gabor filters can be generated using the real part deduced from the complex part (Eq. 1) as:

$$g = \exp\left(-\frac{x'^2 + \gamma^2 y'^2}{2\sigma^2}\right) cos\left(2\pi \frac{x'}{\lambda} + \varphi\right) \quad (2)$$

or by the use of the imaginary part given as:

$$g = \exp\left(-\frac{x'^2 + \gamma^2 y'^2}{2\sigma^2}\right) sin\left(2\pi \frac{x'}{\lambda} + \varphi\right) \quad (3)$$

Value validity for orientation $\theta$ is between 0 and $2\pi$, wavelength $\lambda$ is either $= 2$ or $> 2$, the aspect ratio $\gamma$ ranges between 0 and 1, the phase offset $\varphi$ takes values between $-\pi$ to $\pi$, and the bandwidth $\sigma$, accepts values $> 0$.

### 2.2. Spatial Separable Convolution

Separable convolutions (spatial and depthwise) are more efficient computationally than 2D convolutions. For clarity, an explanation is given using a kernel of size $3 \times 3$ ($height \times width$). Conceptually, a 2D convolution convolves on an image using a $3 \times 3$ kernel for each channel (i.e., 3 $channels$) in a one-stage operation. The depthwise convolution operates in two stages. First, a channel-wise convolution $3 \times 3 \times 1$ ($height \times width \times 1\ channel$) which is then integrated linearly by a point-wise convolution $1 \times 1 \times 3$ ($height \times width \times 3\ channel$). On the other hand, spatial separable convolutions decompose a kernel into two matrices ($3 \times 3\ kernel = [3 \times 1] + [1 \times 3]$) in a one-stage operation to perform convolution on an image. This attribute of spatial separable convolution helps reduce the size and computation used in a DCNN model compared to depthwise convolutions used in the inherited model (MobileNet). However, using spatial convolutions throughout a DCNN architecture may lead to sub-optimal performance. Hence the right blend between convolution types will yield a balanced trade-off between performance and computation. A synopsis of the explanation is tabulated in Table 1.

**Table 1.** Operation complexities of convolutions.

| Conv. type | Kernel and No. of operations | Stages | Complexity |
|---|---|---|---|
| Conv2D | $3 \times 3 \times 3$ = 27 $operations$ | 1 | Elementwise operation increases depending on the number of operations and input shape |
| Spatial separable | $(3 \times 1) + (1 \times 3)$ = 6 $operations$ | 1 | |
| Depthwise separable | $(3 \times 3 \times 1)$ + $(1 \times 1 \times 3)$ = 9 $operations$ | 2 | |

### 2.3. Convolutional Block Attention Module

Recently, improvement in the performances of many state-of-the-art DCNNs has to do with integrating attention mechanisms [18]–[22]. The attention mechanism aids DCNN models on which features to learn and focus on as opposed to non-useful features. Among the variant attention modules, one which stands out is the CBAM due to its lightweight, flexible integration into varying DCNN models and support for end-to-end training. CBAM constitutes a Channel Attention Module (CAM) which guides the DCNN on what features to focus on, and a Spatial Attention Module (SAM) which guides the DCNN on where to search for information in the feature map. Applying CAM and SAM sequentially on feature maps results in refined feature maps that boost the DCNN model's performance. Figure 1 is a graphical representation of the CBAM module.

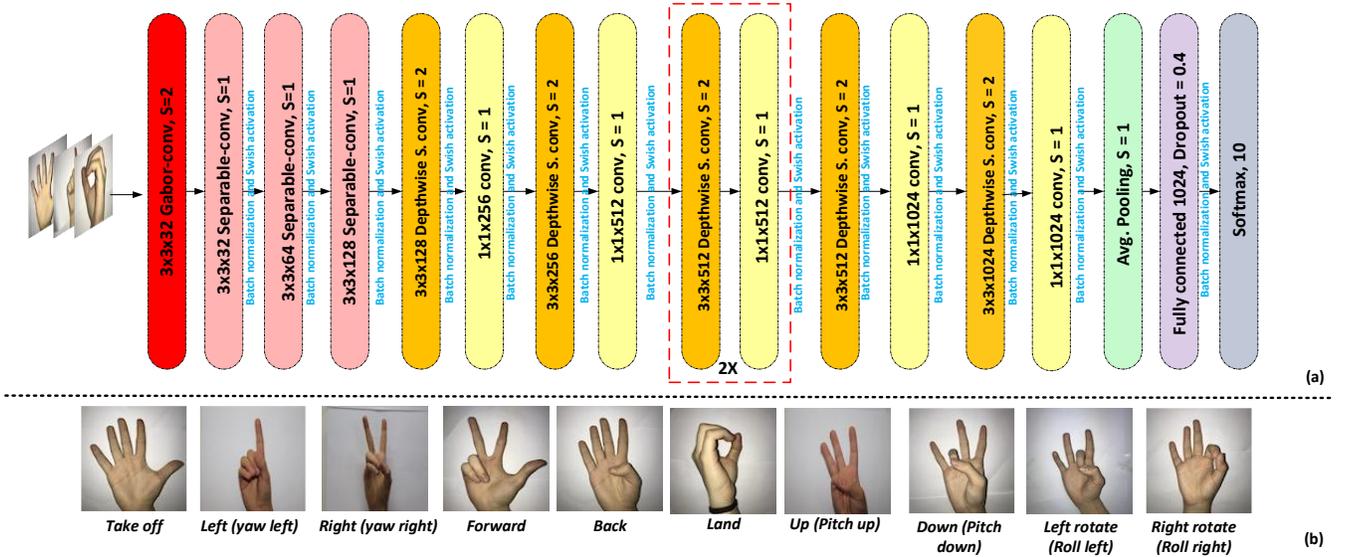

**Fig. 1.** (a) The illustration of G-MobNet architecture, and (b) the representation of hand gesture navigational commands.

### 2.4. G-MobNet Architecture

MobileNet has 28 layers considering depthwise and point-wise convolutions as separate layers and 4.2 million trainable parameters. The MobileNet model starts with a standard convolutional layer before stacked depthwise and point-wise convolutions followed by batch normalization and ReLU activation. An average pooling reduces the spatial convolved features before being fed to a fully connected layer, then a Softmax for classification.

The inherited model (MobileNet) is modified to have fewer trainable parameters hence lesser computation. The 2D convolutional layer is replaced with generic Gabor filters generated in Sec. 2.1. G-MobNet layers are reduced by using spatial separable convolutions in the lower layers of the architecture. The architecture of G-MobNet is further shallowed by reducing the repetitive five layers of output shape (2, 2, 512), as in the base model MobileNet, to 2 layers. The Swish [23] activation function is utilized in the modified model, batch-normalization technique, and drop-out at a probability of 0.4 makes up the model specification. CBAM is integrated after each convolutional layer to refine feature maps. The configuration of G-MobNet was experimentally efficient. A graphical representation of G-MobNet is visualized in Fig. 1a.

### 2.5. Data and Domain Randomization

An open-source hand gesture dataset [24] containing a little of 2000 images forms the basis of the dataset in this study. Data augmentation, which alters images via rotation, flips, hue, and saturation, is performed to increase the dataset to 4150 images. For G-MobNet to learn from various data, the domain randomization technique [25] was introduced. Using Poser software, variant synthetic hand gestures in accord with the gestures of the real data are generated under diverse parameters (e.g., lighting, orientation, textures, etc.). By randomizing the parameters mentioned earlier, G-MobNet is expected to learn robust features from synthetic data and generalize them to the real world. Domain randomization offers enhanced alterations to synthetic data, which aids CNN/DCNN models in learning representations that are relevant to the data. For brevity reasons, Fig. 1b gives an insight into the dataset with the associated corresponding gesture command.

### 2.6. G-MobNet Gesture Controller

To intuitively control the DJI Tello drone, the predictions from G-MobNet are integrated with a perception-based action space and Trajectory Generators herein TG (minimum-jerk trajectory and velocity trajectory). Perception-based action space is utilized to aid the Tello drone in navigating at alternating speeds. At time $t$, a frame from the camera of the Tello drone is divided into $N \times N$ grids, which entails variant speeds (e.g., 2, 4, 6, etc.). Selecting a grid depends on the drone's Field of View (FOV) (i.e., the FOV restricts the search in one out of the four quartiles of the frame) and the prediction accuracy for a particular hand gesture at a threshold of 50%. We weakly restrict the selection space to the accuracy range within which the prediction accuracy falls after the FOV has selected a quartile. A graphical illustration is given in Fig. 2a. For the drone to propagate from $position_x$ to $position_n$, the minimum-jerk trajectory, which optimizes the squared jerk sum along a trajectory, is used together with the velocity trajectory. The minimum-jerk trajectory is given as follows:

$$x(t) = x_i + (x_f - x_i)\left(10\left(\frac{t}{d}\right)^3 - 15\left(\frac{t}{d}\right)^4 + 6\left(\frac{t}{d}\right)^5\right) \quad (4)$$

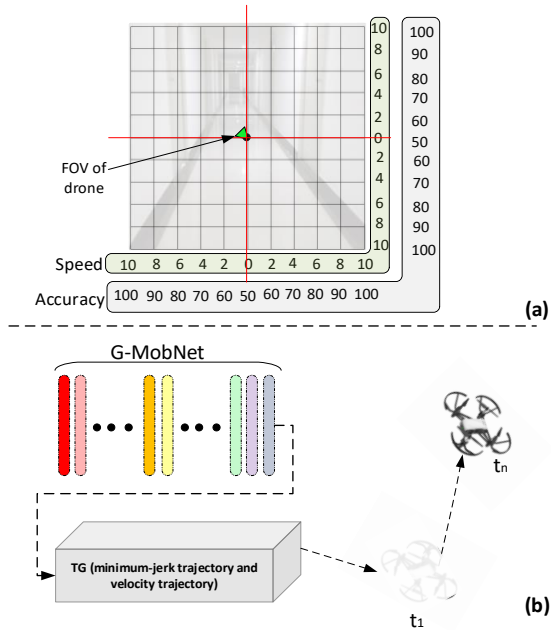

**Fig. 2.** (a) Illustration of probabilistic action space, and (b) G-MobNet gesture controller.

where $x_i$ denotes the current position of the Tello drone, $x_f$ is the new position to move to, $t$ is the travel time, and $d$ is the duration it takes to reach the new position. The velocity trajectory is given as follows:

$$\dot{x}(t) = \frac{1}{d}(x_f - x_i)\left(30\left(\frac{t}{d}\right)^2 - 60\left(\frac{t}{d}\right)^3 + 30\left(\frac{t}{d}\right)^4\right) \quad (5)$$

here the only parameter permissible for alteration is $d$, the duration it should take to get to a final position during locomotion. The other parameters are interpreted as in Eq. 4. The schematic of the G-MobNet gesture controller is visualized in Fig. 2b.

## 3. Experiment and Results

We extensively evaluate our gesture controller (G-MobNet) under variant matrices categorized into two folds; (1) computation performance and (2) flight test performance. Details of the evaluation matrices are given under respective sub-sections. The inherited model (MobileNet) is set as comparator 1 and the second comparator, a conventional model, is a computer vision-based gesture controller based on *MediaPipe module* which provides 3D landmarks for hand poses. The landmark poses (using the landmark IDs) are encoded per the hand gesture commands in Fig. 1b in a Hierarchical Data Format (HDF5) to serve as an artificial intelligence controller model for the computer vision-based approach herein comparator 2. A graphical insight is given in Fig 3.

### 3.1. Computational Performance

Since comparator 2 does not go through a training stage, the computational comparison is between the DCNN models,

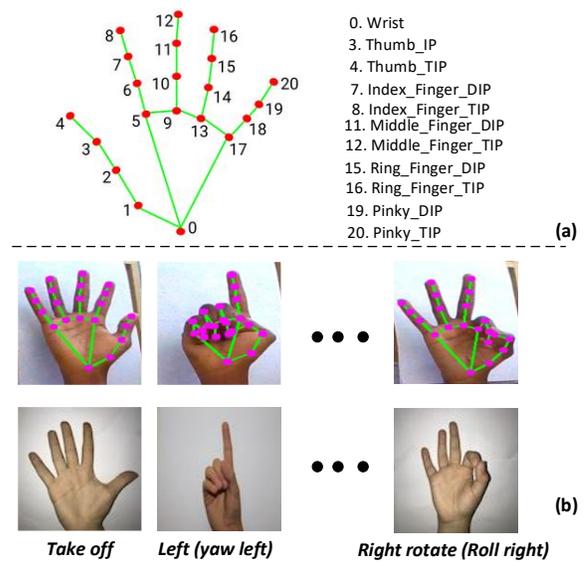

**Fig. 3.** (a) Landmarks for hand poses, and (b) navigational hand gesture commands from landmarks perspective.

G-MobNet, and MobileNet. The implementation was based on the TensorFlow deep learning framework running on an NVIDIA Geforce RTX 2070. The training approach (e.g., batch size, learning rate, etc.) follows the inherited model, MobileNet.

*3.1.1 Computational Consumption*

Using NVIDIA profiling tools (NVVP, NSIGHT, and NVML), numerical values required to compute computational resources consumed by each model during training are adduced. A modified version of computational computation cost [26] is utilized in this study; this is given as:

$$cost = runtime \times gpuload \times gpumem \times data\_cores \quad (6)$$

$$comp.cost = \frac{cost - windows\_process}{training\ epochs} \quad (7)$$

where $runtime$ is the time spent during training, $gpuload$ denotes the voltage received by the GPU cores, $gpumem$ is the memory used, $data\_cores$ denotes the data transferred between the GPU cores and $windows\_process$ which we introduce to account for the windows background process.

From Fig. 4, it can be seen that G-MobNet utilizes lesser computational resources than MobileNet during training, contributing to the overall computational consumption of about 18% less than MobileNet. The computational reduction of G-MobNet is attributed to the configuration of the model. As described in Sec. 2.1, Gabor filters have a computational complexity of $N \times N$ whereas standard convolutions have $(N \times N)^2$. Replacing conventional convolution filters with Gabor filters helps reduce computational resources. In addition, Table 1 shows that using spatial separable convolutions requires three lesser operations than depthwise separable convolutions, which also contributes to the computational reduction in G-MobNet.

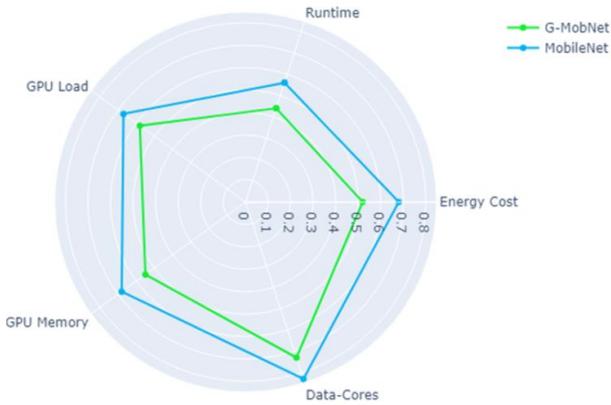

**Fig. 4.** Normalized energy consumption by G-MobNet and MobileNet during training for 100 epochs.

*3.1.2 Accuracy and Loss*

Both models are trained from scratch for 100 iterations using the dataset described in this study. Fig. 5 is a representation of the training and loss accuracies. From Fig. 5a, it can be seen that G-MobNet does not attain any overfitting; hence G-MobNet can generalize to other unseen data. For the case of MobileNet, as seen in Fig. 5b, MobileNet experiences overfitting, limiting the model capabilities of generalizing to unseen data. The tabulated results of the training and testing accuracies from Table 2 show that G-MobNet has a little over 1% accuracy more than MobileNet. We conjecture that using Gabor filters for extracting meaningful features in the lower layers of G-MobNet and introducing DR also contributes to mitigating the overfitting problem. Table 3 summarizes some attributes for both models, which infer, in a way, the computational consumption of each model.

*3.2. Flight Test Performance*

The performance of gesture control of the Tello drone is inferred by trailing two flight trajectories, rectangular and half-rectangular (L-shaped) trajectories. A track displacement, an Extended Kalman filter (EKF) [27], and a quantitative measure, the distance to command response to access the three-hand gesture controllers (G-MobNet, MobileNet, and Comparator 2), are used as evaluation metrics.

*3.2.1 Track displacement*

From Fig. 6a and 6d (rectangular and half-rectangular trajectories), it can be seen that G-MobNet performs better than MobileNet and comparator 2. Besides, G-MobNet navigates smoothly along the trail due to the integration of TG's (minimum-jerk and velocity trajectory generators). Also, perception-based action selection dependent on G-MobNet prediction accuracies aids the drone in navigating intuitively. (i.e., during curves on the trajectory, we observed the choice of selecting speed was mostly low, which fostered smooth navigation). The MobileNet performed better than comparator 2 with minor drifts and hovering behavior at some point in the flight. The controller with the most difficulty is that of comparator 2. Moderate drifts and hovering occurred during the flight test.

**Table 2.** Summary of accuracies and losses.

| Model | Training accuracy | Training loss | Testing accuracy | Testing loss |
|---|---|---|---|---|
| **G-MobNet** | 97.87% | 0.1476 | 96.19% | 0.1647 |
| **MobileNet** | 96.54% | 0.1436 | 86.54% | 0.5236 |

**Table 3.** Synopsis of model attributes.

| Model | No. of layers | Trainable params. | Model size | Avg. epoch compute time |
|---|---|---|---|---|
| G-MobNet | 19 | 3.6m | 27.4MB | 19s |
| MobileNet | 28 | 4.2m | 39.1MB | 30s |

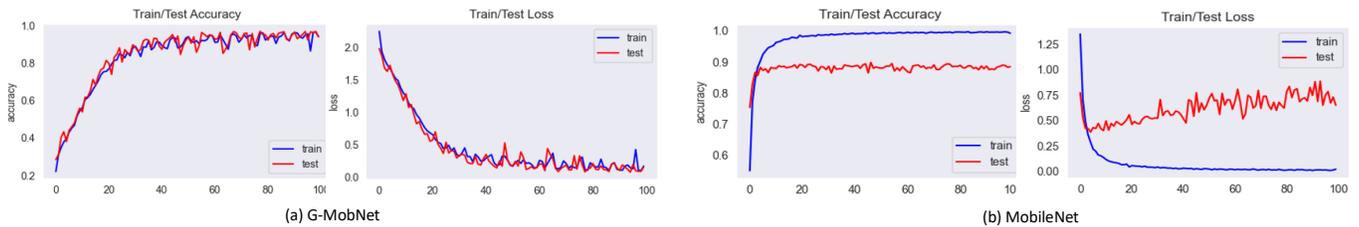

**Fig. 5.** Training and testing accuracies and losses for (a) G-MobNet, and (b)MobileNet.

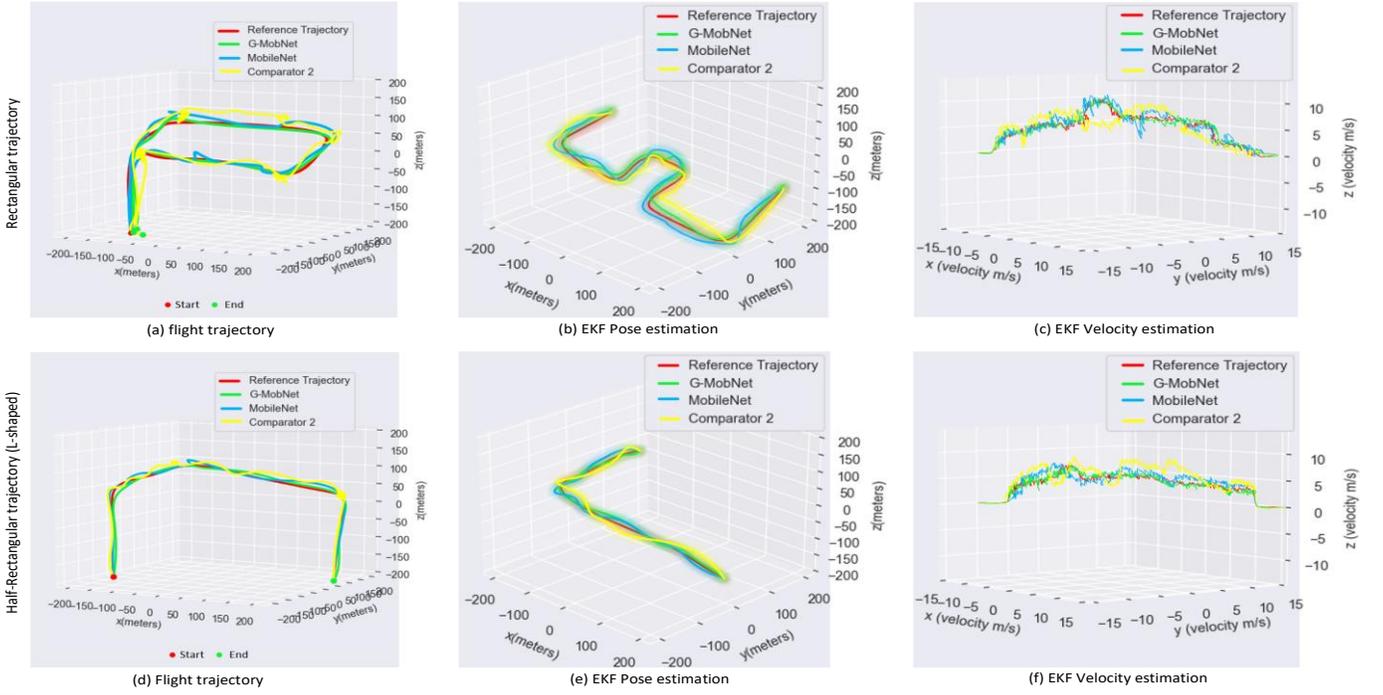

**Fig. 6.** Flight trajectories for rectangular path (a) and half rectangular path (d). Pose estimation via EKF for rectangular trajectory (b) and half rectangular trajectory (e). Velocity estimation using EKF for rectangular trajectory (c) and half rectangular trajectory (f).

*3.2.2 Extended Kalman filter Pose and Velocity Estimation*

The EKF is used as a non-linear state estimator (pose and velocity estimation) for the Tello drone. Using numerical readings from the Inertial Measurement Unit (IMU) of the Tello drone, pose, and velocity estimates can be deduced after deriving the partial derivate of $f(\cdot)$, which represents the previous state of the Tello drone. The function $h(\cdot)$ gives the measurement between the Tello drone's current state $x_k$ and the measurement model $z_k$ which unfolds as:

$$x_k = f(x_{k-1}, u_k) + w_k \qquad (8)$$

$$z_k = h(x_k) + v_k \qquad (9)$$

where $x_{k-1}$ is the Tello drone's previous state and $u_k$ is the input (navigational commands).

***Lemma 1:*** if $w_k$ (state transition model noise) and $v_k$ (measurement model noise) are zero-mean multivariate Gaussian noise for the Tello drone with covariance $Q_k$ and $R_k$, the drones pose, and velocity estimation can be deduced as:

$$F_k = \left.\frac{\partial f}{\partial x}\right|_{\hat{x}_{k-1|k-1}, u_k}, \; H_k = \left.\frac{\partial h}{\partial x}\right|_{\hat{x}_{k|k-1}} \qquad (10)$$

$$\hat{x}_{k|k-1} = f(\hat{x}_{k-1|k-1}, u_k) \qquad (11)$$

$$P_{k|k-1} = F_k P_{k-1|k-1} F_k^T + Q_k \qquad (12)$$

$$\tilde{y}_k = z_k - h(\hat{x}_{k|k-1}) \qquad (13)$$

$$K_k = P_{k|k-1} H_k^T (R_k + H_k P_{k|k-1} H_k^T)^{-1} \qquad (14)$$

$$\hat{x}_{k|k} = \hat{x}_{k|k-1} + K_k \tilde{y}_k \qquad (15)$$

$$P_{k|k} = (I - K_k H_k) P_{k|k-1} \qquad (16)$$

$F_k$ and $H_k$ are the partial derivate of the functions $f(\cdot)$ and $h(\cdot)$. $\hat{x}_{k|k-1}$ denotes predicted state estimation, $P_{k|k-1}$ is the estimated error covariance, $\tilde{y}_k$ denotes measurement residual, $K_k$ is the near-optimal Kalman gain, $\hat{x}_{k|k}$ represents the updated drone state estimate, and $P_{k|k}$ denotes the updated covariance error.

Fig. 6b and 6e denote the EKF pose estimation for the three gesture controllers. It can be seen that G-MobNet can track and follow the reference trajectory with a minimal error rate (maximum absolute error in each component is below 2 meters) which connotes nearly no divergence from the reference trajectories. Among the two comparators, MobileNet performs much better than comparator 2 (computer-vision-based method) but lags behind G-MobNet. The MobileNet model maximum absolute error in each component is under 8 meters, while comparator 2 records 13 meters.

For velocity estimation via EKF, as seen from Fig. 6c and 6f, G-MobNet does not experience spontaneous outbursts in velocities, indicating the IMU readings entail little noise. We conjecture that this results from using TG (velocity trajectory), which streamlines velocities smoothly. MobileNet and comparator 2 periodically show an outburst of velocities since the IMU readings are affected by noise, as seen in Figures 6c and 6f. The continuous reading of such erroneous data causes the velocities estimation of both MobileNet and comparator 2 to diverge from the reference velocity.

*3.2.3 Distance to Command response*

A quantitative inference to measure gesture command response to the Tello drone is tabulated in Table 4 concerning the distance between the human operator and the Tello drone. Two separate environments (clear and mild clear) are used to verify the response rate to gesture commands. As seen from Table 4, G-MobNet and MobileNet attain a competitive response rate of distance to command response. Comparator 2 fails to respond to gesture commands as the distance goes beyond 12 meters. A secondary observation in the second environment (mild clear) is the decrease in response to gesture commands since the environment is perturbed by the lighting condition and the human operator's shadow reflections.

**Table 4.** Summary of distance to command response.

| Model | Environment | Distance to command response |
|---|---|---|
| G-MobNet | | 19.4 meters |
| MobileNet | Clear | 18.8 meters |
| Comparator 2 | | 11.2 meters |
| G-MobNet | | 17.3 meters |
| MobileNet | Mild clear | 15.5 meters |
| Comparator 2 | | 9.3 meters |

## 4. Conclusion

This study reconfigures a state-of-the-art mobile network (MobileNet) by fusing Gabor filters and spatial separable convolutions to reduce the computation requirements during training. The reconfigured model G-MobNet was trained using an open-source dataset combined with synthetic data, which inculcates domain randomization techniques. Prediction commands from G-MobNet were integrated with trajectory generators for smooth navigation. A real-world verification of the presented gesture controller using a low resource-constrained drone (DJI Tello drone) was carried out. Under two main evaluation metrics: (1) computational computation, and (2) flight test performance, G-MobNet exhibited promising results. Under the navigation control of the G-MobNet controller, the DJI Tello drone avoided nearly no drifts, attained smooth navigation, and surpassed MobileNet and comparator 2 controllers, respectively. While G-MobNet has shown promising results, more enhancements can be integrated when scaling to macro drones.

## Acknowledgments

The authors would like to thank Prof. Xiaoling Zhang for her guidance in the study. Further, the authors appreciate Priscilla Fosuah Sakodie and Abigail Boamah for extensive proofreading.